\documentclass[letterpaper]{article} 
\usepackage{aaai2026}  
\usepackage{times}  
\usepackage{helvet}  
\usepackage{courier}  
\usepackage[hyphens]{url}  
\usepackage{graphicx} 
\usepackage{natbib}  
\usepackage{caption} 
\usepackage{algorithm}
\usepackage{algorithmic}

\usepackage{stmaryrd}
\usepackage{newfloat}
\usepackage{listings}
\DeclareCaptionStyle{ruled}{labelfont=normalfont,labelsep=colon,strut=off} 
\floatstyle{ruled}
\newfloat{listing}{tb}{lst}{}
\floatname{listing}{Listing}
\nocopyright 

\title{Towards Numerical TOHTN Planning with SMT-based HTN-SAT Encoding}
\author{
    Takudzwa Togarepi\textsuperscript{\rm 1},
    Gaspard Quenard\textsuperscript{\rm 1},
    Damien Pellier\textsuperscript{\rm 1},
    Humbert Fiorino\textsuperscript{\rm 1}
}
\affiliations{
    \textsuperscript{\rm 1}Univ. Grenoble Alpes, France\\

    \{takudzwa.togarepi, gaspard.quenard, damien.pellier, humbert.fiorino\}@univ-grenoble-alpes.fr
}

\usepackage{amsmath,amsfonts,amssymb}
\newcommand{\name}{\text{\it name}}
\newcommand{\pre}{\text{\it precond}}
\newcommand{\effect}{\text{\it effect}}
\newcommand{\add}{\text{\it effect}^{+}}
\newcommand{\del}{\text{\it effect}^{-}}
\newcommand{\Succ}{\text{\it successor}}

\newtheorem{definition}{Definition}

\begin{document}

\maketitle

\begin{abstract}
While HTN planning has received significant attention in recent years, support for numerical reasoning remains very limited. In this paper, we investigate numerical Totally-Ordered HTN (TOHTN) planning and show how standard SAT-based encodings can be naturally extended with SMT to handle numeric fluents. In addition, we introduce a benchmark suite for numerical TOHTN planning, providing a first common basis for evaluation in this setting. Experimental results show that this simple encoding already constitutes a competitive baseline. This work opens the way to more expressive approaches to HTN planning.
\end{abstract}


\section{Introduction}

Hierarchical Task Network (HTN) planning \cite{erol1994umcp} is a planning paradigm that decomposes complex tasks into simpler subtasks using domain-specific knowledge. Unlike classical planning, HTN introduces abstract tasks, which cannot be executed directly, and methods, which describe how these tasks can be refined into partially ordered sets of subtasks including both primitive tasks (i.e., executable actions) and additional abstract tasks that must themselves be recursively refined. The objective of an HTN planner is to iteratively decompose an initial abstract task into a valid plan (i.e., an executable sequence of primitive tasks). In this paper, we focus our investigation on Totally-Ordered HTN (TOHTN) planning, a highly popular subclass of HTN problems where the decomposition methods specify a totally-ordered list of primitive actions and abstract tasks to be executed in order to achieve an abstract task. 

While HTN planning is a central topic in automated planning and has been included in recent editions of the International Planning Competition (IPC) \cite{behnke2019hierarchical, taitler20242023}, it still suffers from important modeling limitations compared to classical planning formalisms. In particular, numerical and temporal features (such as resource management, costs, and durations) introduced in classical planning with PDDL2.1 \cite{fox2003pddl2} are absent from most HTN planners. Although some work has investigated the integration of temporal aspects into HTN planning \cite{pellier2022hddl}, the formalization of temporal HTN remains an active research topic, and current proposals have not yet reached broad adoption. In contrast, numerical reasoning (e.g., through numerical constraints over the preconditions and effects of primitive tasks) can be naturally incorporated, but has received little attention so far. These features are essential in many real-world applications, including logistics, robotics, and scheduling, where reasoning about quantities is required. To the best of our knowledge, only Siadex \cite{castillo2006efficiently} and Aries \cite{bit2023experimenting} support numerical reasoning in HTN planning. This limitation restricts the applicability of HTN planning to realistic domains.

In this paper, we propose to extend SAT-based TOHTN planning to handle numerical constraints by leveraging Satisfiability Modulo Theories (SMT). SAT-based approaches have shown strong performance in recent years in TOHTN planning due to both the efficiency of modern solvers and improved encodings and search strategies \cite{schreiber2019tree, behnke2018totsat, schreiber2021lilotane, behnke2021block, gaspard2024sibylsat, quenard2025sibylsatopt}, but they are inherently limited to propositional representations. In contrast to heuristic-search-based approaches, which typically require substantial adaptations to handle numerical reasoning, SAT-based methods can be more naturally extended by lifting the encoding to SMT, without fundamentally modifying the search procedure. By moving to SMT, we enable reasoning over numerical variables while preserving the benefits of logical encodings. In this paper, we introduce a new encoding that extends SAT-based TOHTN planning to handle numerical variables and constraints using an SMT solver. Additionally we design and provide seven numerical TOHTN benchmarks to evaluate these approaches. We experimentally show that our SMT-based encoding enables solving numerical TOHTN problems more efficiently than existing numerical HTN planners. 

This paper is organized as follows: first, we introduce the concept of numerical TOHTN planning. Next, we describe the basic incremental encoding used by current SAT-based TOHTN planners to find a solution. Then, we explain how this encoding can be modified to support numerical constraints. Finally, we compare this approach with other numerical HTN planners.

\section{Numerical TOHTN Planning Problem}

We present a formalization of numerical TOHTN planning, building on \cite{behnke2018totsat, behnke2021block, gaspard2024sibylsat} and following the treatment of numeric fluents introduced in PDDL2.1 and discussed for HDDL 2.1 \cite{pellier2022hddl}.

\subsection{Tasks, Actions, Methods, Numeric Fluents, and Task Networks}

Tasks are central to HTN planning. A task is defined by a name and parameters. Tasks are either \textit{primitive} or \textit{abstract}: primitive tasks directly affect the state of the world, while abstract tasks do not; instead, they must be decomposed into primitive tasks using \textit{methods} before they can be executed.

We assume a finite set $L$ of propositions and a finite set $F$ of numeric fluents. A \textit{numeric expression} over $F$ is built from constants and fluents in $F$ using the arithmetic operators $+, -, \times, /$. A \textit{numeric constraint} is an expression of the form $f \bowtie \xi$,  where $f \in F$ is a numeric fluent, $\xi$ is a numeric expression, and $\bowtie \in \{<, \leq, =, \geq, >\}$.

A primitive task $a$ is analogous to an action in classical planning and is defined by a tuple $(\name(a), \pre(a), \effect(a))$. Its preconditions $\pre(a) = (\pre_L(a), \pre_N(a))$ consist of a set of propositional preconditions $\pre_L(a)$ and a set of numeric constraints $\pre_N(a)$. Its effects $\effect(a) = (\add(a), \del(a), \effect_N(a))$ consist of add and delete effects over propositions and a set of numeric effects $\effect_N(a)$. In this work, we restrict numeric effects to assignment effects of the form $f := \xi$, where $f \in F$ and $\xi$ is a numeric expression over $F$.

A state $s$ is defined as a pair $(l,v)$ where $l \subseteq L$ is the set of propositions true in the state, and $v : F \rightarrow \mathbb{R}$ is a valuation function assigning a real value to each numeric fluent. A task $a$ is executable in $s=(l,v)$ iff $\pre_L(a) \subseteq l$ and $v \models \pre_N(a)$, i.e., all numeric constraints in $\pre_N(a)$ are satisfied under $v$. If a task $a$ is executable in a state $s=(l,v)$, applying its effects yields a new state $s'=(l',v')$, where $l'=(l\setminus \del(a))\cup \add(a)$ and $v'$ is obtained by applying $\effect_N(a)$ to $v$.

A method $m=(\name(m),c,w_m)$ indicates how an abstract task $c$ can be refined into a task network $w_m$, called the subtasks of $m$. For notation purposes, we define $M(c) = \{m = (\name(m), c, w_m) \mid m \in M\} $ as the set of all methods that can be applied to decompose the abstract task $c$.

\subsection{Planning Problem and Solution}

\begin{definition}[TOHTN Planning Problem]
A numerical TOHTN planning problem $P$ is a tuple $(L, F, C, A, M, c_I, s_I, g)$ where: $L$ is a finite set of propositions; $F$ is a finite set of numeric fluents; $C$ is a finite set of abstract tasks; $A$ is a finite set of primitive tasks; $M$ is a finite set of decomposition methods; $c_I \in C$ is the initial abstract task to be decomposed; $s_I = (l_I, v_I) \in S$ is the initial state; and $g = (g_L, g_N)$ is the goal condition, where $g_L \subseteq L$ is a set of propositions and $g_N$ is a set of numeric constraints.
\end{definition}

\begin{definition}[TOHTN Planning Problem Solution]
A solution to a numerical TOHTN planning problem $P$ is a primitive task network $\pi \in A^*$ such that:
\begin{enumerate}
    \item $\pi$ is obtained by refining $c_I$ using methods,
    \item $\pi$ is executable in $s_I$,
    \item $\pi$ reaches the goal $g$ after execution.
\end{enumerate}
\end{definition}

\section{SAT-based Search in TOHTN}

\begin{figure}[t]
\centering
\includegraphics[width=.91\columnwidth]{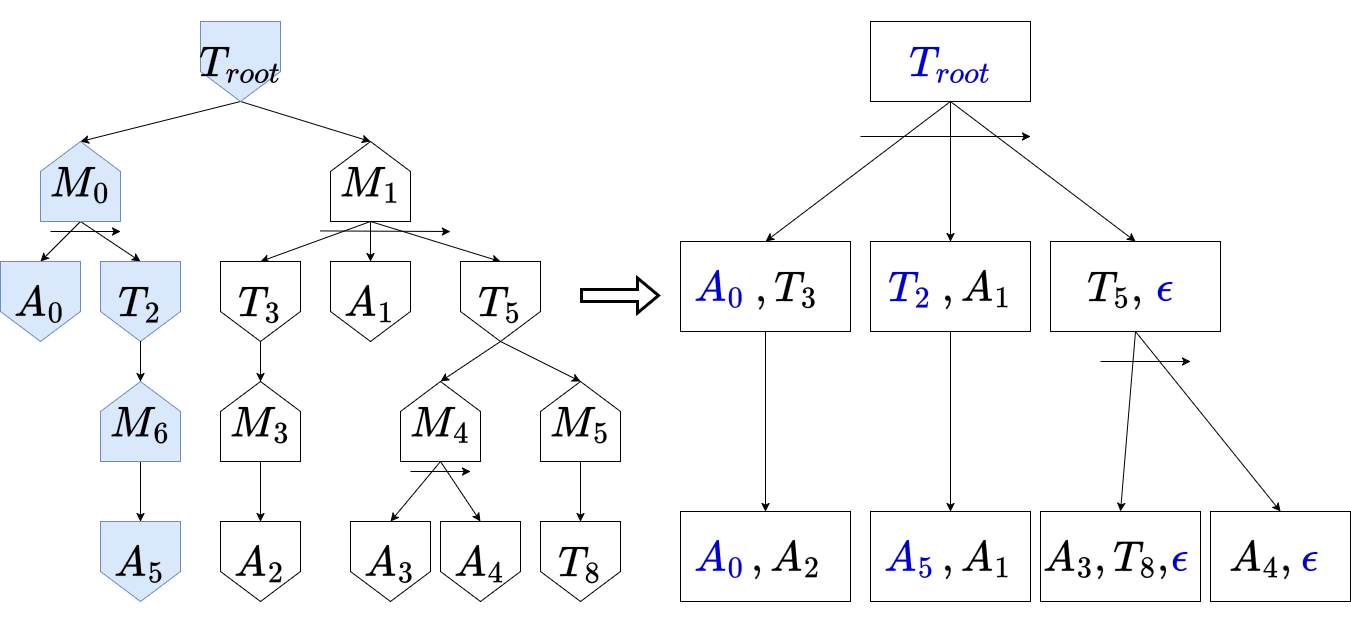} 
\caption{On the left, a simple decomposition schema in the form of an AND/OR tree, where $T_i$ denotes abstract task $i$, $M_i$ denotes method $i$ and $A_i$ denotes action $i$. On the right, we represent the corresponding cPDT. An identical potential solution for both structures is highlighted in blue.}
\label{fig:AND_OR_to_cPDT}
\end{figure}

HTN planning can be naturally represented as an AND/OR tree \cite{ghallab2004automated}, where the root contains the initial abstract task. This tree represents a finite fragment of the potentially infinite decompositions of the initial abstract task, and is incrementally expanded during search. OR nodes correspond to abstract tasks with multiple possible  decompositions (methods), while AND nodes represent methods whose subtasks must all be achieved. A valid plan is obtained by selecting exactly one child for each OR node and all children for each AND node, such that the leaves form a sequence of primitive actions achieving the goal. An example of such AND/OR tree is given in the left side of Figure~\ref{fig:AND_OR_to_cPDT}. This tree is not fully developed here, since the abstract task $T_8$ is undeveloped.


In SAT-based HTN planning, the AND/OR tree is encoded into a SAT formula that is satisfiable iff a solution exists within it, with the satisfying assignment yielding the plan. However, the AND/OR tree alone is insufficient for current encodings. Indeed, clauses must capture action transitions, ensuring preconditions hold at execution time and effects are applied. All current SAT-based HTN planners address this by assigning each task a discrete time step indicating when it may execute; something the AND/OR tree's structural representation does not provide.

That is why all current HTN-SAT planners rely on a structure introduced independently in \textit{TreeRex} and \textit{totSAT} \cite{schreiber2019tree, behnke2018totsat}, which is equivalent to the AND/OR tree while making execution time steps explicit. We call this structure a \emph{Compact  Path Decomposition Tree} (cPDT); its nodes may contain multiple tasks and are organized as follows:
\begin{itemize}
    \item The root node contains only the initial abstract task $c_I$.
    \item To expand a node $P$, its children are built as follows:
    \begin{itemize}
        \item For each abstract task $c$ in $P$ and each method $m_i$ decomposing $c$ into subtasks $\langle t_1, \ldots, t_n \rangle$, the $k$-th child of $P$ contains task $t_k$.
        \item For each primitive task $a$ in $P$, the first child of $P$ contains $a$.
    \end{itemize}
\end{itemize}
The cPDT guarantees that for any task $t$ at leaf $l_i$, all tasks that must precede $t$ appear at leaves $l_j$ with $j < i$, and all that must follow appear at $l_r$ with $r > i$. Consequently, a left-to-right scan of the leaves reveals all possible plans encoded in the cPDT. The cPDT corresponding to the AND/OR tree of Figure~\ref{fig:AND_OR_to_cPDT} is shown on its right side.

We now present an incremental encoding, capturing the core ideas shared by current HTN-SAT encodings \cite{schreiber2019tree, behnke2018totsat, schreiber2021lilotane}, that determines whether a solution exists in a given cPDT. This encoding serves as the basis for the numerical extension introduced next. For each cPDT node $P$, we use Boolean variables $P_t$ indicating that task $t$ is active at $P$, variables $P_p$ indicating that proposition $p \in L$ holds at $P$, and an auxiliary variable $P_{\mathit{prim}}$ indicating that a primitive task is active in $P$. When $t$ is a primitive task, we write it as $a$; when $t$ is an abstract task, we write it as $c$. For any leaf $P$, we denote by $P^+ = \Succ(P)$ the leaf executed immediately after $P$ in the left-to-right order induced by the decomposition hierarchy. To handle the goal state, we introduce a special \emph{virtual} node $G$, representing the goal. Thus, for any node $P$ that has no successor, we define $P^+ = G$. As shown in Figure~\ref{fig:AND_OR_to_cPDT}, there may exist solutions in which no task is active at a given position. To handle this case, we introduce a special action $\varepsilon$ such that $\pre(\varepsilon) = \effect(\varepsilon) = \emptyset$. This action is treated like a normal action in the encoding, but it is ignored in the final plan. We write $P_{\varepsilon}$ to denote that no task is active at node $P$.

\paragraph{Initial task, initial state, and goal.}
At the root node $R$, the initial task and initial state must hold, and the goal must hold at $G$:
\begin{align*}
    &R_{c_I} \\
    &\forall p \in s_I: \ R_p \qquad \forall p \in L \setminus s_I: \ \neg R_p \\
    &\forall p \in g: \ G_p
\end{align*}

\paragraph{Leaf constraints.}
For every leaf node $P$, exactly one task is active at $P$:
\begin{align*}
    &\bigvee_{t \in Tasks(P)} P_t \\
    &\bigwedge_{\substack{t,t' \in Tasks(P) \\ t \neq t'}} \neg(P_t \wedge P_{t'})
\end{align*}
If a primitive task $a$ is active in $P$, then its preconditions must hold at $P$ and its effects at $P^+$:
\begin{align*}
\forall a \in Tasks(P)\cap A:\quad
    &P_a \Rightarrow \bigwedge_{p \in \pre(a)} P_p \\
\forall a \in Tasks(P)\cap A:\quad
    &P_a \Rightarrow \bigwedge_{p \in \add(a)} P^+_p \\
\forall a \in Tasks(P)\cap A:\quad
    &P_a \Rightarrow \bigwedge_{p \in \del(a)} \neg P^+_p
\end{align*}
Frame axioms ensure that if a fact changes value between a node and its successor, it must be explained by a specific action that can change this fact or an abstract task:
\begin{align*}
    &\forall p \in L:\quad \neg P_p \wedge P^+_p \Rightarrow
    \left(\neg P_{\mathit{prim}} \vee \bigvee_{\substack{a \in Tasks(P)\cap A \\ p \in \add(a)}} P_a\right) \\
    &\forall p \in L:\quad P_p \wedge \neg P^+_p \Rightarrow
    \left(\neg P_{\mathit{prim}} \vee \bigvee_{\substack{a \in Tasks(P)\cap A \\ p \in \del(a)}} P_a\right)
\end{align*}
Finally, if an action is chosen on this position, this node is primitive:
\begin{align*}
    &\forall a \in Tasks(P)\cap A:\quad P_a \Rightarrow P_{\mathit{prim}} \\
    &\forall c \in Tasks(P)\cap C:\quad P_c \Rightarrow \neg P_{\mathit{prim}}
\end{align*}

\paragraph{Hierarchical constraints.}
Let $P$ be an expanded node with children $\langle P^1,\dots,P^k \rangle$. Facts hold in $P$ iff they hold in its first child:
\begin{equation*}
    \forall p \in L:\quad P_p \Leftrightarrow P^1_p
\end{equation*}
If an abstract task $c$ is selected at $P$, then exactly one applicable method must be chosen to decompose this task, its subtasks must be assigned to the children in order, and all remaining children must receive $\varepsilon$:
\begin{equation*}
\begin{aligned}
\forall c \in Tasks(P)\cap C:\qquad \\
P_c \Rightarrow \bigvee_{m \in M(c)} \Bigg(
&\bigwedge_{i=1}^{|subtasks(m)|} P^i_{subtasks(m)[i]} \\
&\wedge \bigwedge_{i=|subtasks(m)|+1}^{k} P^i_{\varepsilon}
\Bigg)
\end{aligned}
\end{equation*}
If a primitive task $a$ is selected at $P$ (including $\varepsilon$), it is propagated to the first child and remaining children receive $\varepsilon$:
\begin{equation*}
\forall a \in Tasks(P)\cap A:\quad
P_a \Rightarrow \left(P^1_a \wedge \bigwedge_{i=2}^{k} P^i_{\varepsilon}\right)
\end{equation*}

\paragraph{Primitivity of the leaves.}
Finally, in order to obtain a solution, all leaves must contain primitive actions.
\begin{equation*} \label{eq:encoding_assump_leaves_prim}
    \bigwedge_{P \in leaves(cPDT)} P_{\mathit{prim}}
\end{equation*}

The clauses ensuring the initial task, initial state, and goal are generated only during the first encoding of the cPDT, whereas the clause ensuring the primitivity of the leaves is provided only as \textit{assumptions} (temporary hypotheses active during a single SAT call). All other clauses remain valid after extending the cPDT. When the cPDT is expanded, hierarchical constraints are generated for the newly expanded nodes, and leaf constraints are generated for the new leaf positions (i.e., the children of the expanded nodes).

\subsection{SMT Extension for Numerical TOHTN}

To extend the previous SAT encoding to numerical TOHTN, we introduce for each node $P$ and each numeric fluent $f \in F$ a numeric variable $P_f$ denoting the value of $f$ at $P$. For any numeric expression or constraint $\xi$, we write $\llbracket \xi \rrbracket_P$ for the arithmetic expression or constraint obtained by replacing each fluent $f$ occurring in $\xi$ by the numerical variable $P_f$.

\paragraph{Initial numeric state and numeric goal.}
The initial valuation of the numeric fluents is encoded at the root node:
\[
\forall f \in F:\quad R_f = v_I(f)
\]
Each numeric goal constraint must hold at the goal node $G$:
\[
\forall \xi \in g_N:\quad \llbracket \xi \rrbracket_G
\]

\paragraph{Numeric leaf constraints.}
If a primitive task $a$ is active at leaf node $P$, then its numeric preconditions must hold at $P$:
\[
\forall a \in Tasks(P)\cap A,\ \forall \xi \in \pre_N(a):\quad
P_a \Rightarrow \llbracket \xi \rrbracket_P
\]
If $a$ is active at $P$, then each numeric effect of the form $f = \xi$ determines the value of $f$ at $P^+$:
\[
\forall a \in Tasks(P)\cap A,\ \forall (f = \xi) \in \effect_N(a):
\]
\[
P_a \Rightarrow (P^+_f = \llbracket \xi \rrbracket_P)
\]

\paragraph{Numeric frame axioms.}

For each leaf node $P$ and each fluent $f \in F$, the value of $f$ can change between $P$ and $P^+$ only if the selected task is abstract or a primitive action that can modify $f$:
\[
(P_f \neq P^+_f) \Rightarrow
\left(
\neg P_{\mathit{prim}} \;\vee\;
\bigvee_{\substack{a \in Tasks(P)\cap A \\ f \in \effect_N(a)}} P_a
\right)
\]

\paragraph{Numeric hierarchical constraints.}
As in the propositional case, numeric fluent values are constrained to be equal between a node $P$ and its first child $P^1$:
\begin{equation*}
    \forall f \in F:\quad P_f = P^1_f
\end{equation*}

All Boolean clauses from the SAT encoding remain unchanged. The numeric
clauses described above are generated following the same incremental scheme:
the initial numeric state and numeric goal clauses are generated once at the
first encoding, while the numeric precondition, effect, and frame axiom
clauses are generated for each new leaf node as the cPDT is extended and numerical hierarchical constraints are generated for the newly expanded nodes.

\section{Evaluation}


We compare our approach with the two numerical HTN planners available to us. The evaluation involves three planners: SibylSmt, our planner using the encoding described above with the Z3 SMT solver \cite{de2008z3} and a BFS exploration (i.e., expanding all nodes of the cPDT if no solution is found)\footnote{\url{https://github.com/gaspard-quenard/sibylsat}}; Siadex \cite{castillo2006efficiently}, an HTN planner derived from SHOP \cite{nau1999shop, nau2003shop2} with constraint-based reasoning; and Aries \cite{bit2023experimenting}, a hybrid CP/SAT planner. Experiments were run on an Intel Core i7-12700H with 32GB RAM, with a 10-minute time limit per instance.

To the best of our knowledge, no standard benchmark suite exists for numerical HTN planning. We therefore introduce seven numerical HTN benchmark families, each with an instance generator\footnote{\url{https://github.com/gaspard-quenard/HTN-Numerical-Benchmarks}}. For example, \textsc{Transport-Fuel} models fuel-limited delivery with possible refueling, while \textsc{Transvasement} captures water-jug-style transfers (reaching a target quantity in a specific container by pouring between containers of different capacities).

\begin{table}[ht]
\begin{tabular}{lrlll}
{Domain}& SibylSmt& Aries& Siadex \\
\hline
Alchemy \hfill 20 & 10.54 & 7.79 & \textbf{12.82} \\
Backpack \hfill 20 & \textbf{9.99} & 4.59 & 0 \\
Gripper-colors \hfill 20 & \textbf{9.54} & 7.82 & 6.81 \\
Minecraft \hfill 20 & 9.11 & 4.09 & \textbf{19.91} \\
Overcooked \hfill 20 & \textbf{12.58} & 0 & 1 \\
Transport-Fuel \hfill 20 & \textbf{7.89} & 1.61 & 0 \\
Transvasement \hfill 20 & \textbf{7.17} & 5.82 & 0 \\
\hline
Coverage \hfill 140 & \textbf{91} & 48 & 42 \\
Agile score \hfill 140 & \textbf{66.82} & 31.72 & 40.54 \\
\end{tabular}
\centering
\caption{Performance of each planner on the numerical benchmark suite. Each cell reports the agile score per domain. Total coverage and agile score are shown in the last two rows. Agile score is computed as $Agile\_score=\min\left(1,1-\frac{\log(t)}{\log(T)}\right)$ if a plan is found, and $0$ otherwise, where $T$ is the time limit and $t$ the solving time (in seconds).}
\label{table:overview_results}
\end{table}


Results are shown in Table~\ref{table:overview_results}. Our approach achieves higher coverage and agile scores (reflecting how quickly a planner finds a solution) than the other planners on most benchmarks, suggesting that it is a promising direction for numerical HTN problems. In particular, it appears more robust, being able to solve multiple instances across all benchmark families, while the other planners fail on some domains. Siadex notably struggles on recursive domains with strong combinatorial branching, where the choice of decomposition methods is not easily guided by the current state. 
Finally, none of the planners saturate all the benchmarks, leaving significant room for improvement and the development of more powerful approaches.

\section{Conclusion}

Numerical reasoning is essential in many planning applications, yet remains largely underexplored in HTN planning. We introduced a benchmark suite for numerical TOHTN and a simple SMT-based extension of HTN-SAT encodings to handle numeric fluents. This provides a baseline for future work and enables empirical comparison on shared instances. We hope this work encourages further research on expressive HTN models and stronger planners and benchmarks for numerical hierarchical planning.


\end{document}